\pdfoutput=1

\documentclass[11pt]{article}

\usepackage{emnlp2022}

\usepackage{times}
\usepackage{amsmath}
\usepackage{latexsym}
\usepackage{booktabs}
\usepackage{xcolor}
\usepackage{pgfplots}
\usepackage{tikz}
\usepackage{diagbox}
\usepackage{multirow}
\usepackage{svg}

\pgfplotsset{compat=1.17} 

\usepackage[T1]{fontenc}
\usepackage[utf8]{inputenc}

\usepackage{microtype}

\usepackage{xspace}

\newcommand{\bank}{\textsc{$\langle$BANK$\rangle$}\xspace}
\newcommand{\itech}{\textsc{$\langle$IT$\rangle$}\xspace}
\newcommand{\law}{\textsc{$\langle$LAW$\rangle$}\xspace}
\newcommand{\talk}{\textsc{$\langle$TALK$\rangle$}\xspace}
\newcommand{\relig}{\textsc{$\langle$RELIG$\rangle$}\xspace}
\newcommand{\med}{\textsc{$\langle$MED$\rangle$}\xspace}
\newcommand{\news}{\textsc{$\langle$NEWS$\rangle$}\xspace}
\newcommand{\pad}{\textsc{$\langle$pad$\rangle$}\xspace}

\newcommand{\bleu}{\textsc{BLEU}\xspace}
\newcommand{\comet}{\textsc{COMET}\xspace}

\newcommand{\nmt}{\textsc{NMT}\xspace}
\newcommand{\mdmt}{\textsc{MDMT}\xspace}

\newcommand{\base}{\texttt{base}\xspace}
\newcommand{\ints}{\texttt{ints}\xspace}
\newcommand{\tags}{\texttt{tags}\xspace}

\newcommand{\combined}{\texttt{combined}\xspace}
\newcommand{\mdft}{\texttt{multi-dom\xspace FT}\xspace}
\newcommand{\sdft}{\texttt{single-dom\xspace FT}\xspace}
\newcommand{\indomain}{\texttt{in-dom}\xspace}

\newcommand{\generalbase}{\texttt{general}\xspace\base}
\newcommand{\combinedbase}{\combined \base}
\newcommand{\combinedints}{\combined \ints}
\newcommand{\combinedtags}{\combined \tags}

\newcommand{\idints}{\indomain \ints}
\newcommand{\idtags}{\indomain \tags}

\newcommand{\ftints}{\mdft \ints}
\newcommand{\fttags}{\mdft \tags}

\title{Additive Interventions Yield Robust \\ Multi-Domain Machine Translation Models}

\author{Elijah Rippeth\thanks{\quad Work was done during an internship at Microsoft} \\
  Department of Computer Science \\ University of Maryland \\
  \texttt{erip@cs.umd.edu} \\\And
  Matt Post \\
  Microsoft \\
  \texttt{mattpost@microsoft.com} \\}

\begin{document}
\maketitle
\begin{abstract}
Additive interventions are a recently-proposed mechanism for controlling target-side attributes in neural machine translation.
In contrast to tag-based approaches which manipulate the raw source sequence, interventions work by directly modulating the encoder representation of all tokens in the sequence.
We examine the role of additive interventions in a large-scale multi-domain machine translation setting and compare its performance in various inference scenarios. We find that while the performance difference is small between intervention-based systems and tag-based systems when the domain label matches the test domain, intervention-based systems are robust to label error, making them an attractive choice under label uncertainty. Further, we find that the superiority of single-domain fine-tuning comes under question when training data size is scaled, contradicting previous findings.
\end{abstract}

\section{Introduction}

Multi-domain machine translation (\mdmt) is the paradigm in which a single model is trained to service many domains by training on multiple corpora covering disparate labeled domains. The goal of \mdmt is not only to provide high quality \textit{general} machine translation enabled by knowledge transfer across domains, but also to enable high quality \textit{domain-specific} machine translation when a model is provided cues about the target domain, used to control the generation. Though an intuitive task, the expectations surrounding the task were only recently formalized by \citet{pham-etal-2021-revisiting} in which the authors provided both a set of functional requirements demanded of successful \mdmt models and an experimental framework under which those requirements can be tested.

\citet{pham-etal-2021-revisiting} explored several mechanisms for controlling domain, ranging from simple tag-based approaches to meta-learning based mechanisms. According to the functional requirements outlined by the authors, no method meets all the expectations demanded of effective multi-domain machine translators, though the experiments were run on a relatively small dataset of only in-domain data. The primary remaining expectations, according to the authors, are the superiority of fine-tuning based methods as compared to these methods which can control the target domain, and the ability to accommodate fuzzy or uncertain domains. 

This framework is useful, but the authors leave open several other questions regarding the state of \mdmt. The first of these is data size. Previous experiments focused only on relatively small, in-domain data in an otherwise high-resource setting of English-French and found that most models pale in comparison to models fine-tuned on a single domain. We wonder whether this fine-tuning superiority conclusion holds under a more realistic paradigm in which models trained on large, out-of-domain datasets are fine-tuned on in-domain data. While pretraining and fine-tuning on in-domain data can yield strong in-domain performance---as observed by the authors---this is likely to be at the cost of general domain performance, calling into question the transferability under \mdmt.

Next, we wonder if new methods might help with the issue of domain control in \mdmt. The authors examine reasonable mechanisms for controlling the domain which were known at the time. Since then, new methods have been developed which we hope to investigate under the prescribed framework. We hypothesize that additive interventions \cite{schioppa-etal-2021-controlling}, which learn tag embeddings separately from the encoder, may be harder to ignore, and that the learned interventions may be able to absorb target-side properties more easily, while freeing the encoder to learn strong representations purely for translation. 

In this work we scale the original experimental framework presented in \citet{pham-etal-2021-revisiting} by including a significantly larger, more realistic dataset. We also experiment with additive interventions as an alternative to domain tagging. We find that:

\begin{itemize}
    \item additive interventions perform roughly equivalently with tag-based approaches in the ideal case where provided tags match the target domain.
    \item additive interventions are much more robust in the face of incorrect and uncertain domain labels.
    \item when the experiment is scaled, models fine-tuned targeting a single domain are strong translators, but are never unmatched by other models which can service multiple domains suggesting that \mdmt models in a high-resource setting are competitive with best-in-class baselines.
\end{itemize}

\section{Method}

As a baseline, we inject domain metadata using the tag-based approach. In this scheme, a token representing the target-side attribute, $t$, is prepended to source segment $x$ and fed to the encoder $E$ whose hidden representation is finally exposed to decoder $D$ in a ``normal" fashion:
\begin{align*}
    \hat{y} = D(E(\left[t\right] + x))
\end{align*}
where $+$ indicates sequence concatenation. In tag-based approaches, the expectation is that the domain tag as a prefix acts as a conditioning variable which encourages target-side attributes to appear as desired in the final translation. 

While effective and architecturally non-invasive, this method is not without downsides. Because the target token's contribution to the encoder representation is learned, there is a chance that the attribute can be ignored. To address this and other weaknesses of tag-based approaches, \citet{schioppa-etal-2021-controlling} present the additive interventions method which requires an encoder $E$, a decoder $D$, and a separate attribute embedding layer $Emb$. Given a source segment $x$ and a sentence-level attribute token $t$, we have
\begin{align*}
    V & = Emb(t) \\
    \hat{y} & = D(E(x) \oplus V)
\end{align*}
where $\oplus$ is defined as addition broadcasted along the token dimension. Importantly, this allows prototypically discrete attributes to be represented and controlled in a \textit{continuous} fashion, allowing for interpolation, scaling, and positionally invariant combinations, among other useful features. We note that these are somewhat analogical to an ``additive" version of ``source factors" approaches \cite{hoang-etal-2016-improving, sennrich-haddow-2016-linguistic} with one major difference: additive interventions happen \textit{after} the encoder rather than \textit{before} the encoder.

While the original work only introduces the interventions to the top-most decoder layers in order to allow for partially freezing pretrained networks, we simplify by applying the intervention to the top layer of the encoder, such that it affects all decoder layers. Further, the authors report that improved general performance can be promoted by randomly inducing a zero-vector intervention. As such, we can specify that $t$ is randomly replaced by \pad with some probability with the same effect. We report 20\% masking in this paper, though we experiment with 0\% masking and find no significant differences between the two.

\section{Experimental Setup}

\begin{table}[t]
\centering
\scalebox{0.95}{
\begin{tabular}{lrrrrrrrr}
 \toprule
	Source &	Parallel sents (k) & Source tokens (m)  \\
 \midrule
 ParaCrawl & 229,340 & 4,190.0 \\
 BANK & 190 & 6.3 \\
 IT & 270 & 3.6 \\
 LAW & 501 & 17.1 \\
 TALK & 160 & 3.6 \\
 RELIG & 130 & 3.2 \\
 MED & 2,609 & 133.0 \\
 NEWS & 254 & 5.6 \\
 \bottomrule
  \end{tabular}}
 \caption{Effective training set sizes} \label{tab:data_stats}
\end{table}
\begin{table*}[t]
\centering
\setlength\tabcolsep{2pt}
\scalebox{0.75}{
\begin{tabular}{lrrrrrrrrrrrrrr}
 \toprule
	\multirow{2}{*}{Method} & \multicolumn{2}{c}{BANK} & \multicolumn{2}{c}{IT} & \multicolumn{2}{c}{LAW} & \multicolumn{2}{c}{TALK} & \multicolumn{2}{c}{RELIG} & \multicolumn{2}{c}{MED} & \multicolumn{2}{c}{WMT15} \\
	& \small{\bleu} & \small{\comet} & \small{\bleu} & \small{\comet} &  \small{\bleu} & \small{\comet} &  \small{\bleu} & \small{\comet} &  \small{\bleu} & \small{\comet} &  \small{\bleu} & \small{\comet} & \small{\bleu} & \small{\comet} \\
 \midrule
 \generalbase & 42.4 & 0.485 & 38.3  & 0.311 & 56.2 & 0.832 & 40.6 & 0.585 & 18.9 & 0.166 & 43.9 & 0.548 & 41.3 & 0.639 \\
 \midrule
  \combinedbase & 52.1 & 0.559 & 45.6 & 0.528 & 59.8 & 0.855 & 41.5 & 0.614 & 27.8 & 0.284 & 49.8 & 0.651 & 41.7 & 0.633 \\
\combinedints & 51.9 & \textbf{0.573} & 44.7 & 0.512 & 59.9 & 0.859 & 41.3 & 0.610 & 27.6 & 0.268 & 50.1 & 0.647 & \textbf{41.6} & \textbf{0.638} \\
 \combinedtags & 52.0 & 0.546 & \textbf{46.5} & 0.492 & 59.8 & 0.856 & \textbf{43.7}& \textbf{0.647} & \textbf{28.8} & \textbf{0.307} & 50.1 & 0.647 & 36.8 & 0.606 \\
 \midrule
 \idints & 58.5 & 0.615 & 51.9 & 0.615 & 66.6 & 0.891 & 39.2 & 0.494 & 88.7 & 0.872 & 55.4 & 0.695 & \textbf{30.1} & \textbf{0.289} \\
 \idtags & 58.7 & 0.611 & 51.1 & 0.599 & 66.4 & 0.893 & \textbf{39.8} & \textbf{0.531} & 89.5 & 0.893 & 55.4 & 0.685 & 26.8 & 0.243\\
 \midrule
 \ftints & 56.1 & 0.604 & 50.6 & 0.605 & 64.9 & \textbf{0.896} & 41.3 & 0.580 & 79.4 &0.791 & 51.6 & 0.671 & \textbf{34.3} & 0.433 \\
 \fttags & \textbf{56.9} & 0.614 & 50.9 & 0.595 & 64.8 & 0.870 & 41.6 & \textbf{0.605}  & \textbf{83.6} & \textbf{0.850} & 51.9 & 0.673 & 33.4 & 0.439 \\
 \midrule
 \sdft & 58.2 &0.637 & 50.8 &0.629 & 67.0 &0.917 & 45.1 &0.653 & 39.0 &0.402 & 52.6 & 0.679 & $-$ & $-$ \\
 \bottomrule
 \end{tabular}}
 \caption{MT quality scores per test set. Statistically significant differences between \tags and \ints at the 95\% confidence interval with 1000 bootstrapped samples \textbf{bolded}.} \label{tab:mdmt_results}
\end{table*}
\definecolor{red}{HTML}{d12424}
\definecolor{cyan}{HTML}{34ebc9}
\definecolor{skyblue}{HTML}{34c3eb}
\definecolor{navy}{HTML}{3462eb}
\definecolor{magenta}{HTML}{d934eb}
\definecolor{red}{HTML}{d12424}
\definecolor{cyan}{HTML}{34ebc9}
\definecolor{skyblue}{HTML}{34c3eb}
\definecolor{navy}{HTML}{3462eb}
\definecolor{magenta}{HTML}{d934eb}
\definecolor{pink}{HTML}{9124d1}
\definecolor{lightgreen}{HTML}{abeb34}
\definecolor{darkgreen}{HTML}{31d44f}
\definecolor{orange}{HTML}{e3a127}
\definecolor{gray}{HTML}{adaba8}
\definecolor{yellow}{HTML}{e3c844}

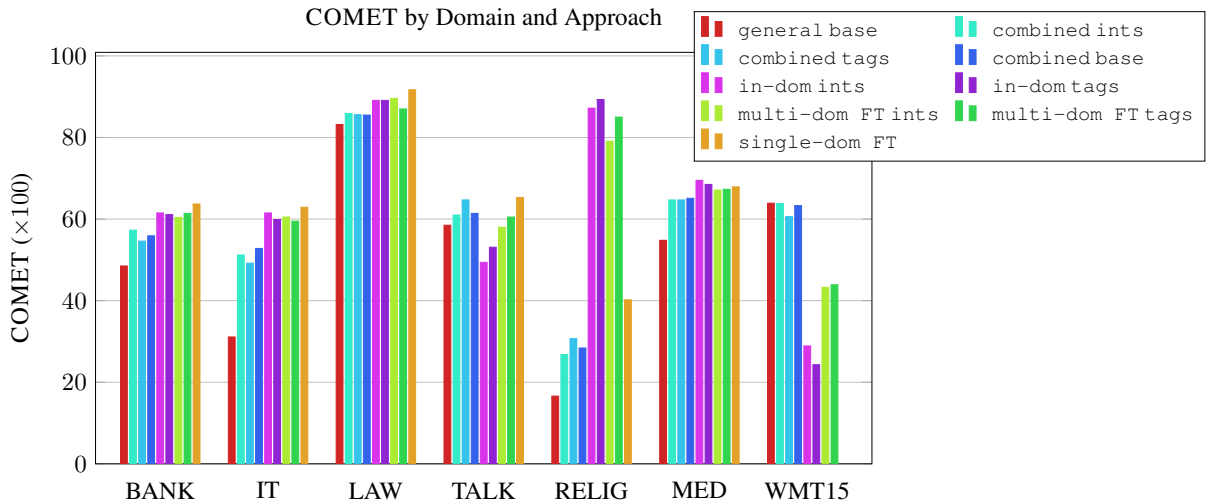
\begin{figure*}[t!]
\centering
\scalebox{0.85}{
\begin{tikzpicture}
    \begin{axis}[
        title = {\comet by Domain and Approach},
        width  = 0.85*\textwidth,
        height = 8cm,
        major x tick style = transparent,
        ybar=1pt, %
        bar width=3pt,
        ymajorgrids = true,
        ylabel = {\comet ($\times 100$)},
        symbolic x coords={BANK,IT,LAW,TALK,RELIG,MED,WMT15},
        xtick = data,
        scaled y ticks = false,
        enlarge x limits=0.1,
        ymin=0,
        legend cell align=left,
        legend columns=2,
        legend style={
                at={(1.1,1.1)},
                anchor=north,
                column sep=1ex
        }
    ]
        \addplot[style={red,fill=red,mark=none}]
            coordinates {(BANK, 48.5) (IT, 31.1) (LAW, 83.2) (TALK, 58.5) (RELIG, 16.6) (MED, 54.8) (WMT15, 63.9)};

        \addplot[style={cyan,fill=cyan,mark=none}]
             coordinates {(BANK,57.3) (IT, 51.2) (LAW, 85.9) (TALK, 61.0) (RELIG, 26.8) (MED, 64.7) (WMT15, 63.8)};

        \addplot[style={skyblue,fill=skyblue,mark=none}]
             coordinates {(BANK, 54.6) (IT,49.2) (LAW,85.6) (TALK, 64.7) (RELIG, 30.7) (MED, 64.7) (WMT15, 60.6)};

        \addplot[style={navy,fill=navy,mark=none}]
             coordinates {(BANK,55.9) (IT,52.8) (LAW,85.5) (TALK,61.4) (RELIG,28.4) (MED,65.1) (WMT15,63.3)};

        \addplot[style={magenta,fill=magenta,mark=none}]
             coordinates {(BANK,61.5) (IT,61.5) (LAW,89.1) (TALK,49.4) (RELIG,87.2) (MED,69.5) (WMT15,28.9)};

        \addplot[style={pink,fill=pink,mark=none}]
             coordinates {(BANK,61.1) (IT,59.9) (LAW,89.1) (TALK,53.1) (RELIG,89.3) (MED,68.5) (WMT15,24.3)};

        \addplot[style={lightgreen,fill=lightgreen,mark=none}]
             coordinates {(BANK,60.4) (IT,60.5) (LAW,89.6) (TALK,58.0) (RELIG,79.1) (MED,67.1) (WMT15,43.3)};

        \addplot[style={darkgreen,fill=darkgreen,mark=none}]
             coordinates {(BANK,61.4) (IT,59.5) (LAW,87.0) (TALK,60.5) (RELIG,85.0) (MED,67.3) (WMT15,43.9)};

        \addplot[style={orange,fill=orange,mark=none}]
             coordinates {(BANK,63.7) (IT,62.9) (LAW,91.7) (TALK,65.3) (RELIG,40.2) (MED,67.9)};

        \legend{\small{\generalbase}, \small{\combinedints}, \small{\combinedtags}, \small{\combinedbase}, \small{\idints}, \small{\idtags}, \small{\ftints}, \small{\fttags}, \small{\sdft}}
    \end{axis}

\end{tikzpicture}}
\caption{\comet scores ($\times 100$) by domain and approach}
\label{chart:comet_by_domain_and_approach}

\end{figure*}

\subsection{Data}

We follow the supervised data settings prescribed by \citet{pham-etal-2021-revisiting} which includes splits from seven domains of varying disparity: BANK, IT, LAW, TALK, RELIG, MED, and NEWS. These domains are drawn from various sources: the European Central Bank corpus (BANK) \cite{tiedemann-2012-parallel}; the documentation for the KDE, Ubuntu, GNOME, and PHP projects from Opus \cite{tiedemann-2009-news} combined to form IT; The JRC-Acquis corpus (LAW) \cite{steinberger-etal-2006-jrc}; TED Talks (TALK) \cite{cettolo-etal-2012-wit3}; the Tanzil translation of the Koran (RELIG); the UFAL Medical corpus v1.0 (MED)\footnote{\url{https://ufal.mff.cuni.cz/ufal_medical_corpus}}; and News Commentary corpus v12 (NEWS) \cite{tiedemann-2012-parallel}. For sake of consistency, we rely on roughly the same splits as provided by the authors,\footnote{\url{https://github.com/qmpham/experiments}}, though we remove duplicates within each domain, which changes the size of each training set slightly. Additionally we include English-French ParaCrawl v9 \cite{banon-etal-2020-paracrawl} to serve as a large out-of-domain training set for some experimental settings. The effective training set sizes are summarized in Table ~\ref{tab:data_stats}.

\subsection{Models}

We consider several models falling into two categories: those trained with (\texttt{control}) and without(\texttt{no control}) a method for selecting the target domain. 

We use approximately the same architecture for all settings, though note that all intervention-based models have an extra embedding layer with the same embedding dimension as the encoder\footnote{Adding $|D|\times1024$ parameters, where $D$ is the set of domain labels}. The basic architecture follows a 12-layer encoder, 6-layer decoder transformer with 8 attention heads each \cite{NIPS2017_3f5ee243}, encoder and decoder feedforward embedding dimensions of 4096, and encoder and decoder embedding dimensions of 1024.

\subsubsection{\texttt{no control}}

We train three models with no training-time information about the domain that the data comes from and, as a consequence, have no ability to explicitly control the target domain:
\begin{enumerate}
    \item we have an out-of-domain baseline which is trained only on ParaCrawl: \generalbase.
    \item we have a model which is trained on the in-domain plus out-of-domain training sets: \combinedbase.
    \item  we have six quasi-oracle fine-tuned models which are produced by fine-tuning the \generalbase model on each target domain's training set; we collectively refer to this set of models as single-domain fine-tuned (\sdft).
\end{enumerate}

\subsubsection{\texttt{control}}

As mechanisms for controlling the target domain we consider:

\begin{enumerate}
    \item prepending the domain tag to the source sequence, \tags
    \item additive interventions with 20\% masking, \ints
\end{enumerate} 

We apply these two methods to three settings: 
\begin{enumerate}
    \item an in-domain plus out-of-domain setting, \combined
    \item an in-domain-only setting, \indomain
    \item a multi-domain fine-tuning setting, \mdft, where \generalbase is fine-tuned on all in-domain data with domain information available at training time.
\end{enumerate}

This results in six models: 

\begin{itemize}
    \setlength\itemsep{0em}
    \item \combined \ints
    \item \indomain \ints
    \item \mdft \ints
    \item \combined \tags
    \item \indomain \tags
    \item \mdft \tags.
\end{itemize}

\subsection{Training}

We train a joint unigram segmentation model \cite{kudo-2018-subword} using SentencePiece \cite{kudo-richardson-2018-sentencepiece} with a vocabulary of size 32k for each setting in \generalbase, \combined, and \indomain (reusing \generalbase's model for \mdft and \sdft). We train each model by sampling 10M sentences randomly, splitting on digits and enabling byte-fallback. We add a special token for each domain for which we have splits: \bank, \itech, \law, \talk, \relig, \med, and \news. We use these models to segment the data as appropriate in each setting.

We use dropout of 0.1 but disable attention dropout and ReLU dropout. We optimize label smoothed cross-entropy loss with a label smoothing factor of 0.1 \cite{7780677} using Adam \cite{Kingma2015AdamAM}. All models are built and trained using fairseq \cite{ott-etal-2019-fairseq}.

For models trained with out-of-domain data, we shard the effective dataset with each shard containing approximately 1b target tokens. For models trained with in-domain data only, we consider the entire combined in-domain dataset to be a single shard. We train for 30 virtual epochs, where a virtual epoch is defined as a single pass over one shard. For models which are fine-tuned, we fine-tune for 10 additional virtual epochs.

Each in-domain training set is assigned a unique special token which is included in the vocabulary and examples drawn from these in-domain training sets are provided the associated special token at training time. Examples from ParaCrawl are assigned no special domain token (i.e., no token is prepended in \tags models and \pad is always provided in \ints models).

\subsection{Evaluation}

We evaluate in three settings to probe various aspects of MT quality:
\begin{itemize}
    \setlength\itemsep{0em}
    \item we evaluate in-domain performance with each model from \texttt{control} and \texttt{no control} to determine the relative effectiveness of the methods of control against methods without control.
    \item we evaluate on the WMT15 English-French test set \cite{bojar-EtAl:2015:WMT} with no domain label provided (i.e., as if the models were in the \texttt{no control} setting) to test catastrophic forgetting \citep{goodfellow-etal-2013-catastrophic} in a general setting. Importantly, while the models trained on in-domain data have been exposed to newswire data, the labels are not provided at test time in this setting. 
    \item we evaluate the effect of providing the incorrect tag to each test set, as computed by SacreBLEU \cite{post-2018-call} and \comet \cite{rei-etal-2020-comet}, to test the resilience of models to label errors
\end{itemize}

\section{Results}

\definecolor{pink}{HTML}{9124d1}
\definecolor{lightgreen}{HTML}{abeb34}
\definecolor{darkgreen}{HTML}{31d44f}
\definecolor{orange}{HTML}{e3a127}

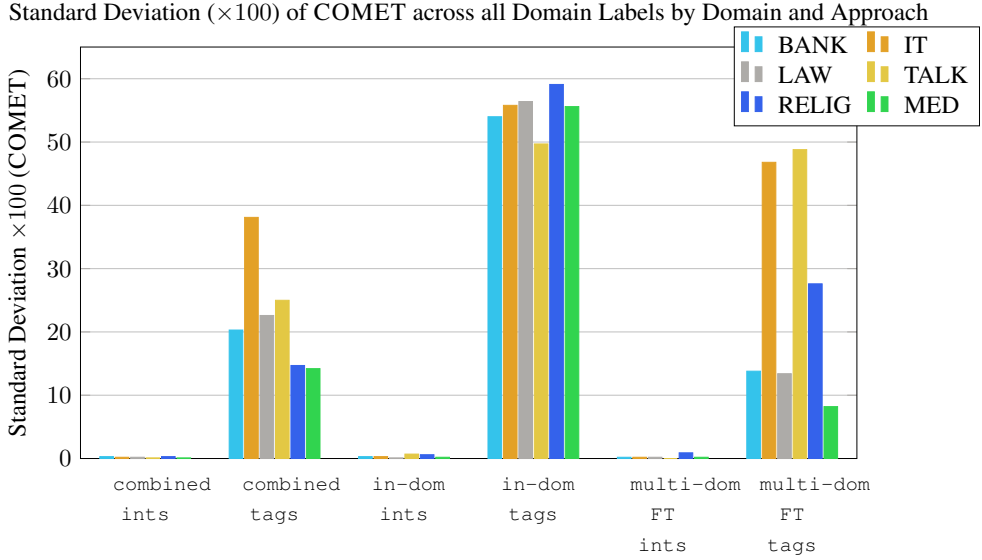
\begin{figure*}[t]
\centering 
\scalebox{0.85}{
\begin{tikzpicture}
    \begin{axis}[
        title = {Standard Deviation ($\times 100$) of \comet across all Domain Labels by Domain and Approach},
        width  = 0.85*\textwidth,
        height = 8cm,
        major x tick style = transparent,
        ybar=2*\pgflinewidth,
        bar width=6pt,
        ymajorgrids = true,
        ylabel = {Standard Deviation $\times 100$ (\comet)},
        symbolic x coords={\small{\combinedints}, \small{\combinedtags}, \small{\idints}, \small{\idtags}, \small{\ftints}, \small{\fttags}},
        xtick = data,
        scaled y ticks = false,
        enlarge x limits=0.1,
        ymin=0,
        legend cell align=left,
        legend columns=2,
        legend style={
                at={(1,1.05)},
                anchor=north,
                column sep=1ex
        },
        xticklabel style={align=center,text width=1cm}
    ]

        \addplot[style={skyblue,fill=skyblue,mark=none}]
             coordinates {(\small{\combinedints},0.3) (\small{\combinedtags},20.3) (\small{\idints},0.3) (\small{\idtags},54.0) (\small{\ftints},0.2) (\small{\fttags},13.8)};

        \addplot[style={orange,fill=orange,mark=none}]
             coordinates {(\small{\combinedints},0.2) (\small{\combinedtags},38.1) (\small{\idints},0.3) (\small{\idtags},55.8) (\small{\ftints},0.2) (\small{\fttags},46.8)};

        \addplot[style={gray,fill=gray,mark=none}]
             coordinates {(\small{\combinedints},0.2) (\small{\combinedtags},22.6) (\small{\idints},0.1) (\small{\idtags},56.4) (\small{\ftints},0.2) (\small{\fttags},13.4)};

        \addplot[style={yellow,fill=yellow,mark=none}]
             coordinates {(\small{\combinedints},0.1) (\small{\combinedtags},25.0) (\small{\idints},0.7) (\small{\idtags},49.7) (\small{\ftints},0.0) (\small{\fttags},48.8)};

        \addplot[style={navy,fill=navy,mark=none}]
             coordinates {(\small{\combinedints},0.3) (\small{\combinedtags},14.7) (\small{\idints},0.6) (\small{\idtags},59.1) (\small{\ftints},0.9) (\small{\fttags},27.6)};
             
        \addplot[style={darkgreen,fill=darkgreen,mark=none}]
             coordinates {(\small{\combinedints},0.1) (\small{\combinedtags},14.2) (\small{\idints},0.2) (\small{\idtags},55.6) (\small{\ftints},0.2) (\small{\fttags},8.2)};
             
        \legend{BANK, IT, LAW, TALK, RELIG, MED}
    \end{axis}
\end{tikzpicture}}
\caption{Impact of domain label error on \comet per test set and approach}
\label{chart:ablate_comet}
\end{figure*}

\begin{figure*}[t!]
    \centering
    \includesvg[scale=0.5]{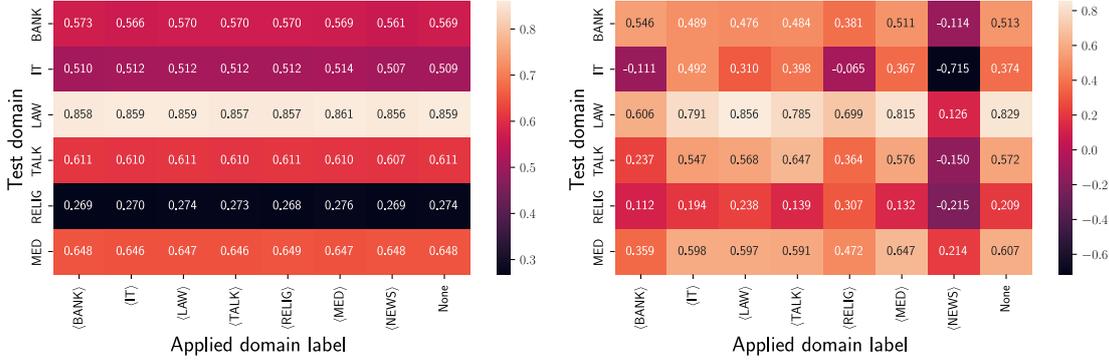}
    \caption{\comet of \combined models under various domain labels. \ints left, \tags right. \ints maintain high quality translations under mismatching domain labels in all cases, unlike \tags.}
    \label{fig:combined_ablation_heatmap}
\end{figure*}
\begin{figure*}[t!]
    \centering
    \includesvg[scale=0.5]{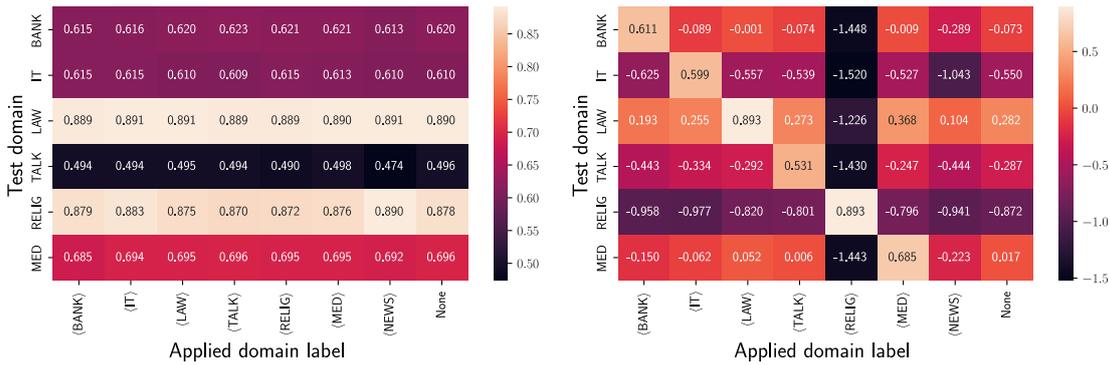}
    \caption{\comet of \indomain models under various domain labels. \ints left, \tags right. \ints maintain high quality translations under mismatching domain labels in all cases, unlike \tags.}
    \label{fig:indomain_ablation_heatmap}
\end{figure*}

\begin{figure*}[t!]
    \centering
    \includesvg[scale=0.5]{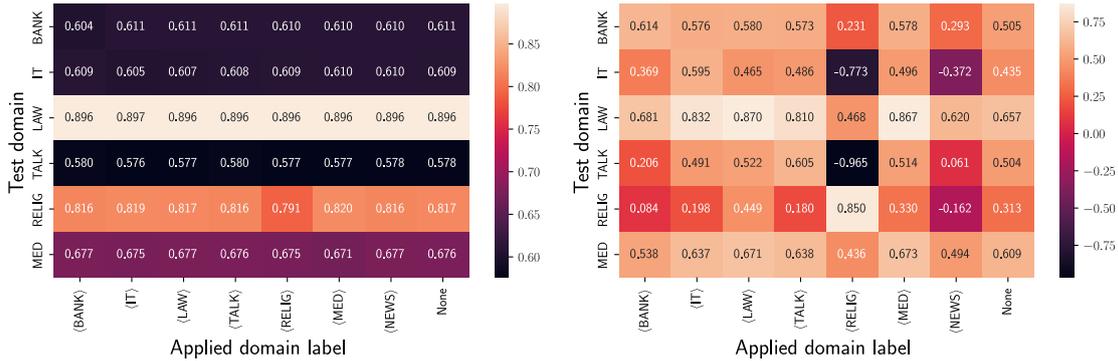}
    \caption{\comet of \mdft models under various domain labels. \ints left, \tags right. \ints maintain high quality translations under mismatching domain labels in all cases, unlike \tags.}
    \label{fig:mdft_ablation_heatmap}
\end{figure*}

\paragraph{No clear winner in ideal case}

We evaluate the setting in which the provided domain label matches the target test domain, and the setting of WMT15 without a provided domain label, for each setting apart from \sdft.
The results can be read in Table~\ref{tab:mdmt_results} and are visualized in Figure~\ref{chart:comet_by_domain_and_approach}. 

Table~\ref{tab:mdmt_results} shows that when comparing \texttt{control} models within a training setting using bootstrap resampling (sample sizes of 1000) \cite{koehn-2004-statistical}, the difference in performance of \tags and \ints are insignificant in the majority of cases. While there are a few cases of statistically significant differences, neither \tags nor \ints are uniformly preferred in these cases. The opposite is observed on the out-of-domain WMT15, where \ints performs uniformly better than \texttt{tags}, often significantly.  

We observe that methods with \texttt{control} in the \combined setting perform approximately equally to the \combinedbase, showing that naive combination of in-domain and out-of-domain with a mechanism to control the domain does not improve over approaches without control, though \indomain and \mdft models tend to perform better on average than any model in the \combined setting.

\paragraph{\ints are robust under domain label mismatch}

Next, we perform an ablation study in which we score each test across all domain label assignments (including the correct label and no label), which allows us to observe the effects of test-time labeling error.
While we compute both \bleu and \comet, we include only \comet here.\footnote{Similar results for \bleu are listed in Appendix~\ref{sec:ablation_results_bleu}}  We include the full results in Tables~\ref{tab:concat_ints_ablate}--\ref{tab:ft_tags_ablate}, but summarize the findings in Figures~\ref{chart:ablate_comet}-\ref{fig:mdft_ablation_heatmap}, which show the robustness of various models and settings to mislabeled domains. 

Figures~\ref{fig:combined_ablation_heatmap}--\ref{fig:mdft_ablation_heatmap} show heatmaps resulting from this ablation, but we refer interested readers to Tables~\ref{tab:concat_ints_ablate}--\ref{tab:ft_tags_ablate} for the long-form charts.  We see that \tags systems' performances vary dramatically, incurring severe degradation in the face of domain label error but performing strongest along the diagonal. \ints systems, on the other hand, see only small performance changes when provided with incorrect domain labels and roughly equal performance under all possible labels, as observed in Figure ~\ref{chart:ablate_comet}. We see that \idtags have the highest average variation in performance, likely owing to the small amount of data which suggests that \idtags overfits to the training data. The variation in performance of \ints systems approaches that of the \generalbase, which by definition ignores the domain label and therefore has 0 variance; however, \ints has demonstrably stronger performance than \generalbase in all domains and, indeed, stronger performance than \tags in a handful of domains and thus seems to learn strong general representations for translation which disentangles the representations of the encoder from the representations of the attribute.

Additionally, through manual analysis we find that \tags systems are more prone to hallucinating translation artifacts from the corpus associated with the domain label being used, often causing quality degradation. We refer to Table~\ref{tab:table120} for an example of such artifacts, which includes topical and target language mismatches along with tokens which appear as a result of the HTML-encoded nature of the \itech dataset.\footnote{Escaping seems to be an artifact of Moses preprocessing leakage of raw data; not germane to all domains in this work.}

\paragraph{Single-domain fine-tuning is not as competitive in large-data settings}

We compare the performance of models trained only with in-domain data and out-of-domain data. From Table~\ref{tab:mdmt_results}, we see slightly stronger in-domain performance for \indomain models as compared to models fine-tuned with out-of-domain data at the cost of out-of-domain performance on WMT15, suggesting that \mdft models generalize better and may surpass \indomain models with more training due to the relatively little fine-tuning budget of 10 epochs afforded to them comparatively.

Finally, we see that while \sdft is typically among the highest performing systems for a given test set, it is never unmatched by an alternative system in \texttt{control}. We observe that \sdft is uniformly stronger than \generalbase and \combined, \indomain and \mdft show competitive in-domain performance. We note that because there is one \sdft model per test set, the effective parameter budget is six times larger than any of the individual models, providing support for both its impracticality and untenability as compared to any other setting. This suggests that single-domain fine-tuning is not as effective as expected in high-resource settings as a strong upper-bound in \mdmt.

\section{Related Work}

Incorporating extra-sentential information has a rich history in \nmt. Aside from controlling for the domain, \citet{sennrich-etal-2016-controlling} use a politeness tag at training and inference time to accommodate coarse politeness control in machine translation. \iffalse Similarly, \citet{rippeth-etal-2022-controlling} use additive interventions when fine-tuning multilingual models to enable cross-lingual transfer of grammatical formality in English-to-X multilingual machine translation. \fi Additionally, \citet{Kuczmarski-Johnson-genderaware} use tags to afford users the ability to vary binary gender in the translations of gender-neutral inputs, hoping to address gender bias in MT. 

At the sub-sequence level, \citet{hoang-etal-2016-improving} and \citet{sennrich-haddow-2016-linguistic} included linguistically-informed word-level ``source factors'', such as part-of-speech tags and dependency relations, as additional feature factors to be concatenated to form a full encoder representation with the goal of reducing ambiguity and sparseness issues.

Perhaps more relatedly, several works have explored the impacts of incorporating domain information into training using various methods. \citet{kobus-etal-2017-domain} explore two methods: a tag-based approach which concatenates a special token to the end of the source sequence, and a ``source factors''-style approach which concatenates domain-level embeddings to each token embedding in the source. \citet{sharaf-etal-2020-meta} explore few-shot domain adaptation, rather than domain control, through the lens of meta-learning and show that a meta-learning based approach is generally stronger than other adaptation approaches, though we note that adaptation and control address different needs. Finally, \citet{stojanovski-fraser-2021-addressing} frame machine translation with document-context as an unsupervised domain adaptation problem and incorporate domain embeddings within the encoder, summed with positional and word embeddings, yielding strong improvements over competitive baseline models.

\section{Conclusion}

In this work we examined the relative impact of additive interventions in a large-scale \mdmt setting. We find that typically there are no significant differences between additive interventions and tag-based approaches when the provided domain label matches the test set, but find that additive interventions exhibit \textit{much more desirable degradation properties} when the domain label is unknown or incorrectly provided. In addition, we find that models first trained on a large, general corpus and then fine-tuned on a single-domain---a realistic baseline in machine translation---rarely perform significantly better than approaches which are trained or fine-tuned only on in-domain data, which is in contrast to their generally superior performance in low-resource settings.

In future work we consider developing extensions to additive interventions which can further improve their performance in \mdmt settings. Additionally, studying additive interventions in other tasks where tag-based approaches are dominant, such as multi-lingual machine translation, could be an interesting avenue for exploration.

\bibliography{anthology,custom}
\bibliographystyle{acl_natbib}

\newpage
\appendix
\onecolumn

\section{Raw scores}
\subsection{Ablation (\comet)}
\begin{table*}[h!]
\centering
\scalebox{0.75}{
\begin{tabular}{|l|l|l|l|l|l|l|l|l|}
\hline
\diagbox{Test set}{Provided label} & \bank & \itech   & \law  & \talk & \relig  & \med  & \news & None \\ \hline
BANK   & 0.573 & 0.566 & 0.570 & 0.570 & 0.570 & 0.569 & 0.561 & 0.569 \\ \hline
IT     & 0.510 & 0.512 & 0.512 & 0.512 & 0.512 & 0.514 & 0.507 & 0.509 \\ \hline
LAW    & 0.858 & 0.859 & 0.859 & 0.857 & 0.857 & 0.861 & 0.856 & 0.859 \\ \hline
TALK   & 0.611 & 0.610 & 0.611 & 0.610 & 0.611 & 0.610 & 0.607 & 0.611 \\ \hline
RELIG  & 0.269 & 0.270 & 0.274 & 0.273 & 0.268 & 0.276 & 0.269 & 0.274 \\ \hline
MED    & 0.648 & 0.646 & 0.647 & 0.646 & 0.649 & 0.647 & 0.648 & 0.648 \\ \hline
\end{tabular}}
\caption{\comet scores of \combinedints under various domain labels}
\label{tab:concat_ints_ablate}
\end{table*}

\begin{table*}[h!]
\centering
\scalebox{0.75}{
\begin{tabular}{|l|l|l|l|l|l|l|l|l|}
\hline
\diagbox{Test set}{Provided label} & \bank & \itech   & \law  & \talk & \relig  & \med  & \news & None \\ \hline
BANK   & 0.546  & 0.489 & 0.476 & 0.484 & 0.381  & 0.511 & -0.114 & 0.513 \\ \hline
IT     & -0.111 & 0.492 & 0.310 & 0.398 & -0.065 & 0.367 & -0.715 & 0.374 \\ \hline
LAW    & 0.606  & 0.791 & 0.856 & 0.785 & 0.699  & 0.815 & 0.126  & 0.829 \\ \hline
TALK   & 0.237  & 0.547 & 0.568 & 0.647 & 0.364  & 0.576 & -0.150 & 0.572 \\ \hline
RELIG  & 0.112  & 0.194 & 0.238 & 0.139 & 0.307  & 0.132 & -0.215 & 0.209 \\ \hline
MED    & 0.359  & 0.598 & 0.597 & 0.591 & 0.472  & 0.647 & 0.214  & 0.607 \\ \hline
\end{tabular}}
\caption{\comet scores of \combinedtags under various domain labels}
\label{tab:concat_tags_ablate}
\end{table*}

\begin{table*}[h!]
\centering
\scalebox{0.75}{
\begin{tabular}{|l|l|l|l|l|l|l|l|l|}
\hline
\diagbox{Test set}{Provided label} & \bank & \itech   & \law  & \talk & \relig  & \med  & \news & None \\ \hline
BANK   & 0.615 & 0.616 & 0.620 & 0.623 & 0.621 & 0.621 & 0.613 & 0.620 \\ \hline
IT     & 0.615 & 0.615 & 0.610 & 0.609 & 0.615 & 0.613 & 0.610 & 0.610 \\ \hline
LAW    & 0.889 & 0.891 & 0.891 & 0.889 & 0.889 & 0.890 & 0.891 & 0.890 \\ \hline
TALK   & 0.494 & 0.494 & 0.495 & 0.494 & 0.490 & 0.498 & 0.474 & 0.496 \\ \hline
RELIG  & 0.879 & 0.883 & 0.875 & 0.870 & 0.872 & 0.876 & 0.890 & 0.878 \\ \hline
MED    & 0.685 & 0.694 & 0.695 & 0.696 & 0.695 & 0.695 & 0.692 & 0.696 \\ \hline
\end{tabular}}
\caption{\comet scores of \idints under various domain labels}
\label{tab:id_ints_ablate}
\end{table*}

\begin{table*}[h!]
\centering
\scalebox{0.75}{
\begin{tabular}{|l|l|l|l|l|l|l|l|l|}
\hline
\diagbox{Test set}{Provided label} & \bank & \itech   & \law  & \talk & \relig  & \med  & \news & None \\ \hline
BANK   & 0.611  & -0.089 & -0.001 & -0.074 & -1.448 & -0.009 & -0.289 & -0.073 \\ \hline
IT     & -0.625 & 0.599  & -0.557 & -0.539 & -1.520 & -0.527 & -1.043 & -0.550 \\ \hline
LAW    & 0.193  & 0.255  & 0.893  & 0.273  & -1.226 & 0.368  & 0.104  & 0.282  \\ \hline
TALK   & -0.443 & -0.334 & -0.292 & 0.531  & -1.430 & -0.247 & -0.444 & -0.287 \\ \hline
RELIG  & -0.958 & -0.977 & -0.820 & -0.801 & 0.893  & -0.796 & -0.941 & -0.872 \\ \hline
MED    & -0.150 & -0.062 & 0.052  & 0.006  & -1.443 & 0.685  & -0.223 & 0.017  \\ \hline
\end{tabular}}
\caption{\comet scores of \idtags under various domain labels}
\label{tab:id_tags_ablate}
\end{table*}

\begin{table*}[h!]
\centering
\scalebox{0.75}{
\begin{tabular}{|l|l|l|l|l|l|l|l|l|}
\hline
\diagbox{Test set}{Provided label} & \bank & \itech   & \law  & \talk & \relig  & \med  & \news & None \\ \hline
BANK   & 0.604 & 0.611 & 0.611 & 0.611 & 0.610 & 0.610 & 0.610 & 0.611 \\ \hline
IT     & 0.609 & 0.605 & 0.607 & 0.608 & 0.609 & 0.610 & 0.610 & 0.609 \\ \hline
LAW    & 0.896 & 0.897 & 0.896 & 0.896 & 0.896 & 0.896 & 0.896 & 0.896 \\ \hline
TALK   & 0.580 & 0.576 & 0.577 & 0.580 & 0.577 & 0.577 & 0.578 & 0.578 \\ \hline
RELIG  & 0.816 & 0.819 & 0.817 & 0.816 & 0.791 & 0.820 & 0.816 & 0.817 \\ \hline
MED    & 0.677 & 0.675 & 0.677 & 0.676 & 0.675 & 0.671 & 0.677 & 0.676 \\ \hline
\end{tabular}}
\caption{\comet scores of \ftints under various domain labels}
\label{tab:ft_ints_ablate}
\end{table*}

\begin{table*}[h!]
\centering
\scalebox{0.75}{
\begin{tabular}{|l|l|l|l|l|l|l|l|l|}
\hline
\diagbox{Test set}{Provided label} & \bank & \itech   & \law  & \talk & \relig  & \med  & \news & None \\ \hline
BANK   & 0.614 & 0.576 & 0.580 & 0.573 & 0.231  & 0.578 & 0.293  & 0.505 \\ \hline
IT     & 0.369 & 0.595 & 0.465 & 0.486 & -0.773 & 0.496 & -0.372 & 0.435 \\ \hline
LAW    & 0.681 & 0.832 & 0.870 & 0.810 & 0.468  & 0.867 & 0.620  & 0.657 \\ \hline
TALK   & 0.206 & 0.491 & 0.522 & 0.605 & -0.965 & 0.514 & 0.061  & 0.504 \\ \hline
RELIG  & 0.084 & 0.198 & 0.449 & 0.180 & 0.850  & 0.330 & -0.162 & 0.313 \\ \hline
MED    & 0.538 & 0.637 & 0.671 & 0.638 & 0.436  & 0.673 & 0.494  & 0.609 \\ \hline
\end{tabular}}
\caption{\comet scores of \fttags under various domain labels}
\label{tab:ft_tags_ablate}
\end{table*}

\subsection{Ablation (\bleu)}
All scores reported are from SacreBLEU\footnote{{\texttt{BLEU|nrefs:1|case:mixed|eff:no|tok:13a|smooth:exp|version:2.2.0}}} \cite{post-2018-call}.

\label{sec:ablation_results_bleu}
\begin{table}[h!]
\centering
\scalebox{0.75}{
\begin{tabular}{|l|l|l|l|l|l|l|l|l|}
\hline
\diagbox{Test set}{Provided label} & \bank & \itech   & \law  & \talk & \relig  & \med  & \news & None \\ \hline
BANK   & 51.9 & 51.7 & 51.9 & 51.9 & 51.9 & 51.8 & 51.8 & 51.9 \\ \hline
IT     & 44.6 & 44.7 & 44.8 & 44.8 & 44.6 & 44.7 & 44.7 & 44.6 \\ \hline
LAW    & 59.8 & 59.8 & 59.9 & 59.8 & 59.7 & 59.8 & 59.7 & 59.9 \\ \hline
TALK   & 41.3 & 41.3 & 41.4 & 41.3 & 41.4 & 41.3 & 41.1 & 41.5 \\ \hline
RELIG  & 27.6 & 27.8 & 27.7 & 27.8 & 27.6 & 27.9 & 27.5 & 27.7 \\ \hline
MED    & 50.0 & 50.0 & 50.0 & 50.0 & 49.9 & 50.1 & 50.0 & 50.0 \\ \hline
\end{tabular}}
\caption{\bleu scores of \combinedints under various domain labels}
\end{table}

\begin{table}[h!]
\centering
\scalebox{0.75}{
\begin{tabular}{|l|l|l|l|l|l|l|l|l|}
\hline
\diagbox{Test set}{Provided label} & \bank & \itech   & \law  & \talk & \relig  & \med  & \news & None \\ \hline
BANK   & 52.0 & 43.5 & 43.0 & 40.4 & 39.0  & 45.2 & 30.2 & 44.2 \\ \hline
IT     & 18.5 & 46.5 & 36.3 & 39.9 & 26.5 & 37.2 & 11.0  & 35.0 \\ \hline
LAW    & 50.2 & 56.4 & 59.8 & 50.7 & 51.4 & 55.5 & 36.9 & 56.2 \\ \hline
TALK   & 29.5 & 39.2 & 38.1 & 43.7 & 28.3 & 39.7 & 22.7 & 37.1 \\ \hline
RELIG  & 21.6 & 24.4 & 25.5 & 16.3 & 28.8 & 18.9 & 14.5 & 22.6 \\ \hline
MED    & 43.5 & 48.5 & 48.3 & 47.3 & 45.0 & 50.1 & 41.6 & 49.1 \\ \hline
\end{tabular}}
\caption{\bleu scores of \combinedtags under various domain labels}
\end{table}

\begin{table*}[h!]
\centering
\scalebox{0.75}{
\begin{tabular}{|l|l|l|l|l|l|l|l|l|}
\hline
\diagbox{Test set}{Provided label} & \bank & \itech   & \law  & \talk & \relig  & \med  & \news & None \\ \hline
BANK   & 58.5 & 58.6 & 58.6 & 58.6 & 58.5 & 58.8 & 58.2 & 58.7 \\ \hline
IT     & 52.0 & 51.9 & 51.4 & 51.4 & 51.8 & 51.6 & 51.4 & 51.8 \\ \hline
LAW    & 66.1 & 66.2 & 66.1 & 66.0 & 65.9 & 66.1 & 66.0 & 66.1 \\ \hline
TALK   & 39.0 & 39.1 & 39.1 & 39.2 & 39.1 & 39.2 & 38.8 & 39.0 \\ \hline
RELIG  & 89.2 & 89.2 & 89.0 & 88.7 & 88.7 & 89.2 & 89.3 & 89.1 \\ \hline
MED    & 55.4 & 55.5 & 55.3 & 55.4 & 55.4 & 55.4 & 55.4 & 55.5 \\ \hline
\end{tabular}}
\caption{\bleu scores of \idints under various domain labels}
\end{table*}

\begin{table*}[h!]
\centering
\scalebox{0.75}{
\begin{tabular}{|l|l|l|l|l|l|l|l|l|}
\hline
\diagbox{Test set}{Provided label} & \bank & \itech   & \law  & \talk & \relig  & \med  & \news & None \\ \hline
BANK   & 58.7 & 31.2 & 36.0 & 34.3 & 3.9  & 36.1 & 27.3 & 34.4 \\ \hline
IT     & 15.5 & 51.1 & 16.6 & 20.0 & 0.4  & 18.8 & 5.9  & 15.9 \\ \hline
LAW    & 42.2 & 43.5 & 66.4 & 45.3 & 12.4 & 48.2 & 40.2 & 44.7 \\ \hline
TALK   & 18.6 & 21.0 & 20.7 & 39.8 & 1.0  & 23.8 & 17.2 & 21.5 \\ \hline
RELIG  & 6.2  & 6.1  & 8.2  & 8.7  & 89.5 & 9.0  & 5.5  & 7.6  \\ \hline
MED    & 32.2 & 33.2 & 32.8 & 33.3 & 5.5  & 55.4 & 29.5 & 33.5 \\ \hline
\end{tabular}}
\caption{\bleu scores of \idtags under various domain labels}
\end{table*}

\begin{table*}[h!]
\centering
\scalebox{0.75}{
\begin{tabular}{|l|l|l|l|l|l|l|l|l|}
\hline
\diagbox{Test set}{Provided label} & \bank & \itech   & \law  & \talk & \relig  & \med  & \news & None \\ \hline
BANK   & 56.1 & 55.9 & 56.5 & 56   & 56.1 & 56.4 & 55.3 & 56.3 \\ \hline
IT     & 50.6 & 50.6 & 50.0 & 50.4 & 50.3 & 50.6 & 49.8 & 50.9 \\ \hline
LAW    & 64.8 & 64.7 & 64.9 & 64.9 & 64.8 & 65.2 & 64.5 & 65.0 \\ \hline
TALK   & 41.2 & 40.8 & 41.1 & 41.3 & 41.3 & 41.2 & 40.4 & 41.5 \\ \hline
RELIG  & 80.4 & 81.1 & 80.5 & 80.2 & 79.4 & 81.8 & 79.3 & 82.2 \\ \hline
MED    & 51.7 & 51.3 & 51.6 & 51.7 & 51.7 & 51.6 & 51.3 & 51.7 \\ \hline
\end{tabular}}
\caption{\bleu scores of \ftints under various domain labels}
\end{table*}

\begin{table*}[h!]
\centering
\scalebox{0.75}{
\begin{tabular}{|l|l|l|l|l|l|l|l|l|}
\hline
\diagbox{Test set}{Provided label} & \bank & \itech   & \law  & \talk & \relig  & \med  & \news & None \\ \hline
BANK   & 56.9 & 54.5 & 54.4 & 52.0 & 49.6 & 55.0 & 43.4 & 54.9 \\ \hline
IT     & 43.1 & 50.9 & 47.4 & 46.9 & 28.0 & 46.9 & 17.3 & 40.8 \\ \hline
LAW    & 55.7 & 63.7 & 64.8 & 61.2 & 59.4 & 64.2 & 55.9 & 60.3 \\ \hline
TALK   & 28.0 & 37.4 & 36.1 & 41.6 & 8.4  & 36.1 & 23.1 & 36.2 \\ \hline
RELIG  & 32.6 & 38.6 & 61.9 & 22.9 & 83.6 & 50.7 & 19.2 & 49.1 \\ \hline
MED    & 49.6 & 51.4 & 51.8 & 50.4 & 49.4 & 51.9 & 49.7 & 51.2 \\ \hline
\end{tabular}}
\caption{\bleu scores of \fttags under various domain labels}
\end{table*}

\newpage
\section{Figures (\bleu)}

\label{sec:results_bleu}
\definecolor{red}{HTML}{d12424}
\definecolor{cyan}{HTML}{34ebc9}
\definecolor{skyblue}{HTML}{34c3eb}
\definecolor{navy}{HTML}{3462eb}
\definecolor{magenta}{HTML}{d934eb}
\definecolor{red}{HTML}{d12424}
\definecolor{cyan}{HTML}{34ebc9}
\definecolor{skyblue}{HTML}{34c3eb}
\definecolor{navy}{HTML}{3462eb}
\definecolor{magenta}{HTML}{d934eb}
\definecolor{pink}{HTML}{9124d1}
\definecolor{lightgreen}{HTML}{abeb34}
\definecolor{darkgreen}{HTML}{31d44f}
\definecolor{orange}{HTML}{e3a127}
\definecolor{gray}{HTML}{adaba8}
\definecolor{yellow}{HTML}{e3c844}

\begin{figure}[h!]
\centering
\scalebox{0.75}{
\begin{tikzpicture}
    \begin{axis}[
        title = {\bleu by Domain and Approach},
        width  = 0.85*\textwidth,
        height = 8cm,
        major x tick style = transparent,
        ybar=1pt, %
        bar width=3pt,
        ymajorgrids = true,
        ylabel = {\bleu},
        symbolic x coords={BANK,IT,LAW,TALK,RELIG,MED,WMT15},
        xtick = data,
        scaled y ticks = false,
        enlarge x limits=0.1,
        ymin=0,
        legend cell align=left,
        legend columns=2,
        legend style={
                at={(1.05,1.0)},
                anchor=north,
                column sep=1ex
        }
    ]
        \addplot[style={red,fill=red,mark=none}]
            coordinates {(BANK, 42.4) (IT, 38.3) (LAW, 56.2) (TALK, 40.6) (RELIG, 18.9) (MED, 43.9) (WMT15, 41.3)};

        \addplot[style={cyan,fill=cyan,mark=none}]
             coordinates {(BANK, 51.9) (IT, 44.7) (LAW, 59.9) (TALK, 41.3) (RELIG, 27.6) (MED, 50.1) (WMT15, 41.6)};

        \addplot[style={skyblue,fill=skyblue,mark=none}]
             coordinates {(BANK, 52.0) (IT,46.5) (LAW,59.8) (TALK, 43.7) (RELIG, 28.8) (MED, 50.1) (WMT15, 36.8)};

        \addplot[style={navy,fill=navy,mark=none}]
             coordinates {(BANK,52.1) (IT,45.6) (LAW,59.8) (TALK,41.5) (RELIG,27.8) (MED,49.8) (WMT15,41.7)};

        \addplot[style={magenta,fill=magenta,mark=none}]
             coordinates {(BANK,58.5) (IT,51.9) (LAW,66.6) (TALK,39.2) (RELIG,88.7) (MED,55.4) (WMT15,30.1)};

        \addplot[style={pink,fill=pink,mark=none}]
             coordinates {(BANK,58.7) (IT,51.1) (LAW,66.4) (TALK,39.8) (RELIG,89.5) (MED,55.4) (WMT15,26.8)};

        \addplot[style={lightgreen,fill=lightgreen,mark=none}]
             coordinates {(BANK,56.1) (IT,50.6) (LAW,64.9) (TALK,41.3) (RELIG,79.4) (MED,51.6) (WMT15,34.3)};

        \addplot[style={darkgreen,fill=darkgreen,mark=none}]
             coordinates {(BANK,56.9) (IT,50.9) (LAW,64.8) (TALK,41.6) (RELIG,83.6) (MED,51.9) (WMT15,33.4)};

        \addplot[style={orange,fill=orange,mark=none}]
             coordinates {(BANK,58.2) (IT,50.8) (LAW,67.0) (TALK,45.1) (RELIG,39.0) (MED,52.6)};

        \legend{\small{\generalbase}, \small{\combinedints}, \small{\combinedtags}, \small{\combinedbase}, \small{\idints}, \small{\idtags}, \small{\ftints}, \small{\fttags}, \small{\sdft}}
    \end{axis}

\end{tikzpicture}}
\caption{\bleu scores by domain and approach}
\label{chart:bleu_by_domain_and_approach}

\end{figure}
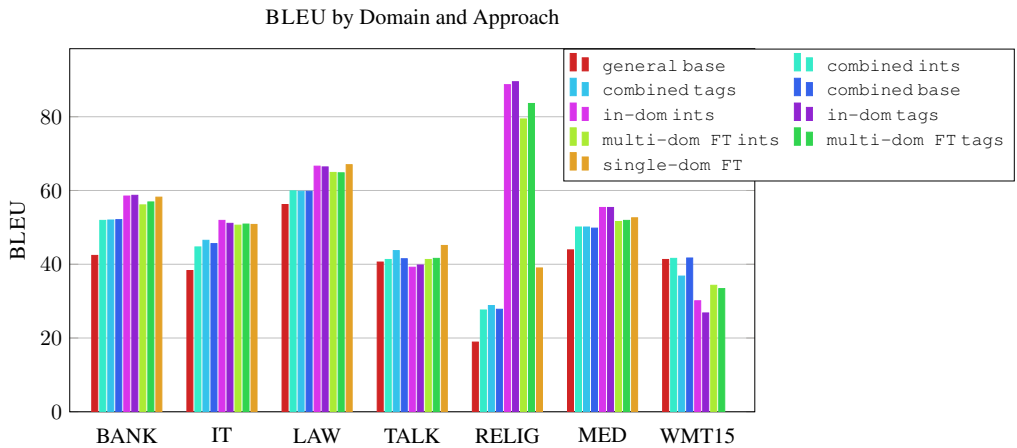

\definecolor{pink}{HTML}{9124d1}
\definecolor{lightgreen}{HTML}{abeb34}
\definecolor{darkgreen}{HTML}{31d44f}
\definecolor{orange}{HTML}{e3a127}

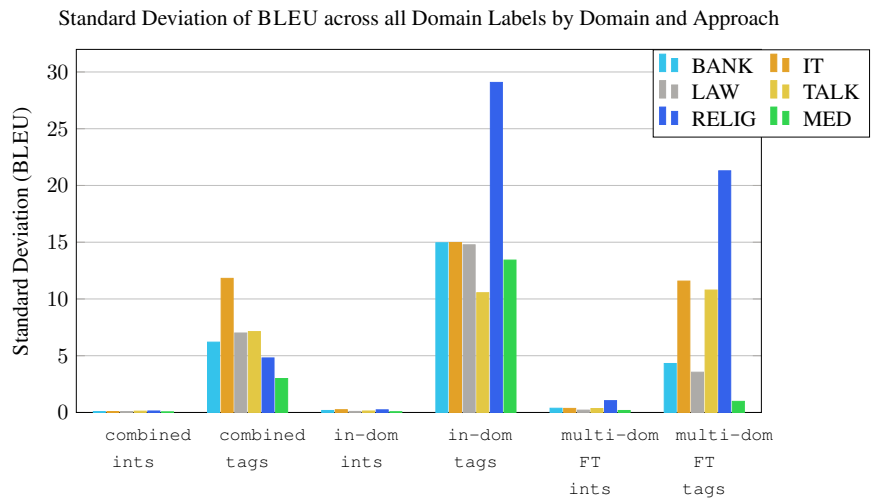
\begin{figure*}[h!]
\centering
\scalebox{0.75}{
\begin{tikzpicture}
    \begin{axis}[
        title = {Standard Deviation of \bleu across all Domain Labels by Domain and Approach},
        width  = 0.85*\textwidth,
        height = 8cm,
        major x tick style = transparent,
        ybar=2*\pgflinewidth,
        bar width=6pt,
        ymajorgrids = true,
        ylabel = {Standard Deviation (\bleu)},
        symbolic x coords={\small{\combinedints}, \small{\combinedtags}, \small{\idints}, \small{\idtags}, \small{\ftints}, \small{\fttags}},
        xtick = data,
        scaled y ticks = false,
        enlarge x limits=0.1,
        ymin=0,
        legend cell align=left,
        legend columns=2,
        legend style={
                at={(1,1)},
                anchor=north,
                column sep=1ex
        },
        xticklabel style={align=center,text width=1cm}
    ]

        \addplot[style={skyblue,fill=skyblue,mark=none}]
             coordinates {(\small{\combinedints},0.075593) (\small{\combinedtags},6.197335) (\small{\idints},0.176777) (\small{\idtags},14.95106) (\small{\ftints},0.37321) (\small{\fttags},4.315897)};

        \addplot[style={orange,fill=orange,mark=none}]
             coordinates {(\small{\combinedints},0.083452) (\small{\combinedtags},11.81669) (\small{\idints},0.244584) (\small{\idtags},14.97157) (\small{\ftints},0.358569) (\small{\fttags},11.57657)};

        \addplot[style={gray,fill=gray,mark=none}]
             coordinates {(\small{\combinedints},0.075593) (\small{\combinedtags},7.002436) (\small{\idints},0.091613) (\small{\idtags},14.77314) (\small{\ftints},0.20702) (\small{\fttags},3.549245)};

        \addplot[style={yellow,fill=yellow,mark=none}]
             coordinates {(\small{\combinedints},0.116496) (\small{\combinedtags},7.126497) (\small{\idints},0.130247) (\small{\idtags},10.55191) (\small{\ftints},0.34641) (\small{\fttags},10.78623)};

        \addplot[style={navy,fill=navy,mark=none}]
             coordinates {(\small{\combinedints},0.130931) (\small{\combinedtags},4.805874) (\small{\idints},0.232993) (\small{\idtags},29.08058) (\small{\ftints},1.041205) (\small{\fttags},21.29183)};
             
        \addplot[style={darkgreen,fill=darkgreen,mark=none}]
             coordinates {(\small{\combinedints},0.053452) (\small{\combinedtags},2.990819) (\small{\idints},0.064087) (\small{\idtags},13.42745) (\small{\ftints},0.175255) (\small{\fttags},0.979705)};
             
        \legend{BANK, IT, LAW, TALK, RELIG, MED}
    \end{axis}
\end{tikzpicture}}
\caption{Impact of domain label error on \bleu per test set and approach}
\label{chart:ablate_bleu}
\end{figure*}

\newpage
\section{Examples}
\begin{table}[h!]
  \centering
  \scalebox{0.85}{
  \begin{tabular}{c|cc} \hline
Src & Never; soon they will deny ever worshipping them, and will turn into their opponents. &  \\ \hline
Ref & Bien au contraire! {[}ces divinités{]} renieront leur adoration et seront pour eux des adversaires. &  \\ \hline
\ftints & Bien au contraire! {[}ces divinités{]} renieront leur adoration et seront pour eux des adversaires. \\ \hline
\fttags & You are about to translate the 'None 'COMMAND, there are \\ & some rules on how to translate it. Please see http: / / / /www.mysql.com /.  \\ \hline\hline
Src & And the evil-doers say: Ye are but following a man bewitched.  \\ \hline
Ref & Les injustes disent: «Vous ne suivez qu'un homme ensorcelé».  \\ \hline
\idints & Les injustes disent: «Vous ne suivez qu'un homme ensorcelé». \\ \hline
\idtags & Et les « \& \#160; diaboliques \& \#160; » disent \& \#160;: « \& \#160; fired \& \#160; » \\ &  est le suivant d'un homme.  \\
  \end{tabular}}
  \caption{Example translation artifacts from incorrect domain label; a translation of \relig sentences with \itech domain label under different models. We note that the HTML-encoded artifact ``\& \#160;" appears with high frequency in \itech.}
  \label{tab:table120}
\end{table}

\end{document}